\documentclass[10pt, conference]{IEEEtran}
\IEEEoverridecommandlockouts
\usepackage{cite}
\usepackage{amsmath,amssymb,amsfonts}
\usepackage{algorithmic}
\usepackage{graphicx}
\usepackage{textcomp}
\usepackage{xcolor}

\usepackage{color}
\usepackage{subcaption}
\usepackage{hyperref}
\usepackage{booktabs}

\usepackage{amsthm}
\newtheoremstyle{boldremark}
    {\dimexpr\topsep/2\relax} % space above
    {\dimexpr\topsep/2\relax} % space below
    {\itshape}          % body font
    {}          % indent amount
    {\bfseries} % theorem head font
    {:}         % punctuation after theorem head
    {.5em}      % space after theorem head
    {}          % theorem hed spec. (empty = "normal")

\theoremstyle{boldremark}
\newtheorem*{remark}{Remark}

\def\BibTeX{{\rm B\kern-.05em{\sc i\kern-.025em b}\kern-.08em
    T\kern-.1667em\lower.7ex\hbox{E}\kern-.125emX}}
\begin{document}

\title{Self-supervised Spatial-Temporal Learner for Precipitation Nowcasting\\
% {\footnotesize \textsuperscript{*}Note: Sub-titles are not captured for https://ieeexplore.ieee.org  and should not be used}
\thanks{$^{\dagger}$ Corresponding author}
}

\author{\IEEEauthorblockN{Haotian Li$^{\dagger}$, Arno Siebes, Siamak Mehrkanoon}
\IEEEauthorblockA{\textit{Department of Information and Computing Science, Utrecht University, Utrecht, The Netherlands} \\
\{h.li2, a.p.j.m.siebes, s.mehrkanoon\}@uu.nl}

% Anonymous
% \author{\IEEEauthorblockN{Author 1, Author 2, Author 3}

% \and
% \IEEEauthorblockN{4\textsuperscript{th} Given Name Surname}
% \IEEEauthorblockA{\textit{dept. name of organization (of Aff.)} \\
% \textit{name of organization (of Aff.)}\\
% City, Country \\
% email address or ORCID}
% \and
% \IEEEauthorblockN{5\textsuperscript{th} Given Name Surname}
% \IEEEauthorblockA{\textit{dept. name of organization (of Aff.)} \\
% \textit{name of organization (of Aff.)}\\
% City, Country \\
% email address or ORCID}
% \and
% \IEEEauthorblockN{6\textsuperscript{th} Given Name Surname}
% \IEEEauthorblockA{\textit{dept. name of organization (of Aff.)} \\
% \textit{name of organization (of Aff.)}\\
% City, Country \\
% email address or ORCID}
}

\maketitle

\begin{abstract}
Nowcasting, the short-term prediction of weather, is essential for making timely and weather-dependent decisions. 
Specifically, precipitation nowcasting aims to predict precipitation at a local level within a 6-hour time frame. 
This task can be framed as a spatial-temporal sequence forecasting problem, where deep learning methods have been particularly effective. 
However, despite advancements in self-supervised learning, most successful methods for nowcasting remain fully supervised. 
Self-supervised learning is advantageous for pretraining models to learn representations without requiring extensive labeled data. 
In this work, we leverage the benefits of self-supervised learning and integrate it with spatial-temporal learning to develop a novel model, SpaT-SparK. 
SpaT-SparK comprises a CNN-based encoder-decoder structure pretrained with a masked image modeling (MIM) task and a translation network that captures temporal relationships among past and future precipitation maps in downstream tasks.
We conducted experiments on the NL-50 dataset to evaluate the performance of SpaT-SparK. 
The results demonstrate that SpaT-SparK outperforms existing baseline supervised models, such as SmaAt-UNet, providing more accurate nowcasting predictions.
\end{abstract}


\section{Introduction}\label{sec:intro}

Nowcasting is the short-term prediction of weather and is crucial for weather-dependent decision-making. The World Meteorological Organization defines nowcasting as forecasting with local detail, by any method, over a period from the present to six hours ahead, including a detailed description of the present weather \cite{wmoguideline}. 
Nowcasting plays a crucial role in weather-dependent activities, for instance: 
(1) accurate nowcasting addresses the challenge of flood prevention posed by sudden precipitation or extremes \cite{Imhoff2020SpatialAT}; 
(2) it allows water resource management to optimize reservoir levels, plan irrigation schedules, and prevent shortages \cite{Bea2023OptimizingRW}; 
(3) nowcasting informs cities’ decisions in managing stormwater drainage systems, prevent urban flooding, and ensure the safety of residents \cite{foresti2016development}.

Given the cruciality of nowcasting, numerical weather prediction (NWP) models have been developed \cite{johnson1922weather} to forecast the weather. 
NWP models are meteorology-theory-guided mathematical models for weather forecasting and, as a result, involve complex equations representing the dynamics and thermodynamics of the atmosphere \cite{browning1989nowcasting}. 
However, NWP models pose challenges in nowcasting tasks. Solving such equations is non-trivial and time-consuming, leading to the delayed outcomes of NWP models \cite{bech2012doppler}. Therefore, NWP models struggle to meet the “short-time ahead” requirement for nowcasting. 

% ML/DL in nowcasting

% Examples ML/DL in nowcasting
Machine learning and deep learning models are promising candidates for nowcasting/forecasting applications \cite{shi2015convolutional, agrawal2019machine, trebing2021smaatunet, ravuriSkillfulPrecipitationNowcasting2021,mehrkanoon2019deep,trebing2020wind}. 
% Advantages of ML/DL method
Extensively trained, these models achieve fast inference. 
% The foundation? What makes it possible?
In addition, the development of these models benefits from the increasing availability of radar image data, due to advances in remote sensing techniques. 
Trained with the precipitation data, CNN-based models \cite{agrawal2019machine, trebing2021smaatunet} demonstrated accurate performances. 
% spatial-temporal learning
Precipitation nowcasting problem can be formulated as a spatial-temporal sequence problem as well, both the input and prediction target are spatial-temporal sequences \cite{shi2015convolutional}.

% What is SSL?
Self-supervised learning (SSL) is a label-free learning paradigm. 
In SSL, the model is trained on pretext tasks using the data itself to generate supervisory signals, rather than relying on external labels provided by humans. 
% Why choose SSL?
SSL has shown its power in learning representations in various application domains such as computer vision \cite{kolesnikov2019revisiting}, biomedical signal analysis \cite{banville2021uncovering, kazatzidis2024novel}, natural language processing \cite{elnaggar2021prottrans} among others. 
%SSL can enhance the performance of learning model accuracy, robustness, uncertainty \cite{chen2020big, hendrycks2019using}. 
%\cite{pathak2016context} did the first attempt to learn image representations with masked image modeling (MIM). 
%MIM enables learning by masking out part of the image and then reconstructing it, and the model is optimized by minimizing the reconstruction error. MIM is widely adopted in the area of SSL for learning image representations for various downstream tasks \cite{he2022masked, xie2022simmim, gupta2023siamese, bachmann2022multimae}. 
SSL improves model accuracy, robustness, and uncertainty \cite{chen2020big, hendrycks2019using}. \cite{pathak2016context} pioneered masked image modeling (MIM), which masks part of an image and reconstructs it to minimize error. MIM is widely used in SSL for learning image representations in various tasks \cite{he2022masked,gupta2023siamese,bachmann2022multimae}.

% Existing work MAE has shown this point. 
Since SSL is proved to be beneficial to learning representations, we design a self-supervised learning pipeline to help the model extract meaningful representations from the precipitation map data and boost the performance of precipitation nowcasting. 
With the success of enabling SSL on CNNs with SparK\cite{tian2023designing}, we propose the \textit{\textbf{\color{red} Spa}}tial-\textit{\textbf{\color{red} T}}emporal \textit{\textbf{\color{red}SparK}} model (termed as SpaT-SparK) for precipitation nowcasting.
In summary, the contribution of this work is two-fold: 
\begin{itemize}
  \item Providing a simple self-supervised spatial-temporal learning model, SpaT-SparK, for precipitation nowcasting task.
  \item Validating the effect of proposed SpaT-SparK on the precipitation data in the Netherlands. 
\end{itemize}
% \begin{figure}[h] % fig:visualization
%   \centering
%   \includegraphics[width=\columnwidth]{example-image-a}
%   \caption{SpaT-SparK precipitation predictions v.s. ground truths.} \label{fig:visualization_secintro}
% \end{figure}

% High-level illustration
% TODO: finish the high-level illustration
The SpaT-SparK model comprises two parts to perform the precipitation nowcasting task: 
(1) an encoder-decoder structure, which is pretrained with the MIM task, learning to encode and decode the representation of input precipitation maps; 
(2) a translation network to capture temporal relationships between the representation of the past and that of the future precipitation maps. 
In the self-supervised pretraining MIM task, the encoder learns how to encode a sequence of precipitation maps to the latent representation, and the decoder learns how to reconstruct the original maps. 
Subsequently, the translation network learns to capture the temporal dependency between representations of past and future precipitation maps in downstream tasks. The code is available at \url{https://github.com/htlee6/SpaT-SparK}

\section{Related Work}

\textbf{Deep Learning for Precipitation Nowcasting:}
Deep learning is deployed to tackle precipitation nowcasting tasks. 
ConvLSTM \cite{shi2015convolutional} leveraged the temporal dependency among each image input dynamically and improved the performance by extending fully connected LSTM to convolutional structures. 
\cite{agrawal2019machine} used UNet \cite{ronneberger2015u}, a 2D convolutional encoder-decoder architecture, to capture the structural difference between different image inputs to make predictions. 
SmaAt-UNet\cite{trebing2021smaatunet}, AA-TransUNet\cite{yang2022aatransunet}, SAR-UNet\cite{renault2023sarunet} and Broad-UNet \cite{fernandez2021broad} further developed variations of UNet tailored for nowcasting tasks. 
%SmaAt-UNet reduced the model’s parameter size and achieved comparable performance in single frame precipitation nowcasting task by introducing depth-wise separable convolution (DSC) \cite{chollet2017xception} and convolutional block attention module (CBAM) \cite{woo2018cbam} in UNet structures. AA-TransUNet and SAR-UNet furthered improved the performance with the TranUNet backbone and residual DSC blocks, respectively. The Broad-UNet, with asymmetric parallel convolutions and an ASPP module, captures complex patterns by merging multi-scale features while using fewer parameters than the original UNet. Besides the deterministic models mentioned above, \cite{ravuriSkillfulPrecipitationNowcasting2021} utilized the power of generative probabilistic models for precipitation nowcasting tasks and predicted reliably. The authors in \cite{reulen2024ga} introduced GA-SmaAt-GNet model, a novel generative adversarial framework for extreme precipitation nowcasting. \cite{vatamany2024gd} formulated the precipitation nowcasting as a spatialtemporal graph sequence nowcasting problem.
SmaAt-UNet reduced parameter size and achieved comparable performance in precipitation nowcasting by introducing depth-wise separable convolutions (DSC) \cite{chollet2017xception} and CBAM \cite{woo2018cbam} in UNet structures. AA-TransUNet and SAR-UNet further improved performance using the TranUNet backbone and residual DSC blocks, respectively. The Broad-UNet, with asymmetric parallel convolutions and an ASPP module, captures complex patterns with fewer parameters than UNet. Beyond deterministic models, \cite{ravuriSkillfulPrecipitationNowcasting2021} employed generative probabilistic models for reliable predictions, while \cite{reulen2024ga} introduced GA-SmaAt-GNet, a generative adversarial framework for extreme precipitation nowcasting. \cite{vatamany2024gd} framed the task as spatial-temporal graph sequence nowcasting. Large-scale models were developed in the weather forecasting/nowcasting task. 
GraphCast \cite{lam2023learning}, Pangu-Weather \cite{bi2023accurate}, FourCastNet \cite{kurth2023fourcastnet} were large deep learning models trained on large scale data and achieving optimal performances. 

\textbf{Self-supervised Learning:} 
SSL is a promising way to signal supervisory by designing pretext tasks. 
SSL is categorized as contrastive, generative, and contrastive-generative \cite{liuSSLGenerative2021,gui2023survey}. There are masked language modeling (MLM) and masked image modeling (MIM) for learning representations in language and image tasks respectively.
As an instance of MLM, BERT \cite{devlin2018bert} uses a unique $\mathtt{[MASK]}$ token at random positions to enable self-supervised pretraining on large-scale unlabeled corpora. 
Successive works, e.g. RoBERTa \cite{liu2019roberta} were built upon the findings of BERT. 
Masked auto-encoder (MAE) \cite{he2022masked} extended the idea of MLM from natural language to computer vision and learned featured latent representations from image data with a ViT-based \cite{dosovitskiy2020image} encoder and decoder. 
Motivated by MAE, Man et al. proposed W-MAE \cite{man2023wmae}, which utilizes MAE in multi-variable weather forecasting and precipitation forecasting tasks on a 6-hour basis (ERA5 \cite{hersbach2020era5} dataset). 
SparK \cite{tian2023designing} breaks through the limitation of specific ViT-based encoder-decoders, and generalize MIM to CNN-based encoders-decoders with the successfully verified benefits of hierarchical features in vision tasks. % TODO: rephrase
SimVP v2 \cite{tan2022simvp} reveals the relations and differences between self-supervised learning and spatial-temporal predictive learning, and provides a recurrent-free model example \cite{tan2023openstl}.  % TODO: to be finished
Despite these developments, the performance of self-supervised weather forecasting model, such as W-MAE, is tested on long-interval dataset. A self-supervised nowcasting model is necessary for the prediction in the near future. 
In this work, we introduce SpaT-SparK, a self-supervised spatial-temporal learning model motivated by SparK and SimVP v2, to tackle the short-time lead precipitation nowcasting problem on the Netherlands precipitation dataset.

\section{Approach}

\begin{figure*}[htbp] % figure schematic model
  \centering
  \includegraphics[width=0.92\linewidth]{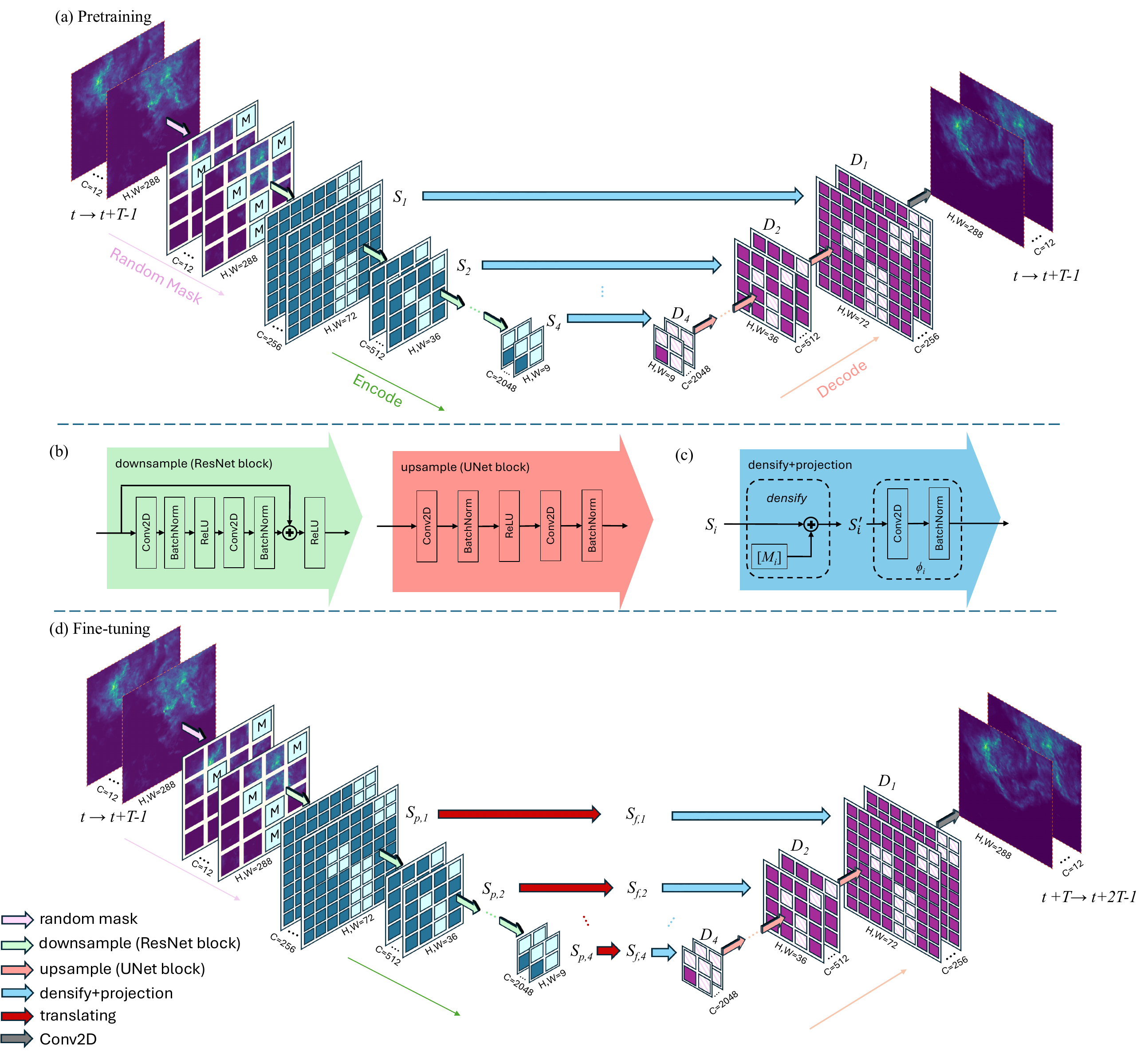}
  \caption{Schematic of SpaT-SparK model in pretraining and fine-tuning mode. 
  (a) In pretraining, only the encoder, the decoder, and the densifying network are pretrained. 
  (b) Architecture of encoder (ResNet) blocks and decoder (UNet) blocks.
  (c) Illustration of densify network and successive projection.
  (d) In fine-tuning, the encoder, the decoder, and the densifying network are initialized with pretrained weight, and the translation network is trained from scratch. The 4th hierarchy is adapted for better visualization purpose; the visualizations of the outputs in (a) and (d) are adapted for illustration purposes, and do not reflect the actual predictions.}
  \label{fig:model_schematic}
  \end{figure*}

The model schematic is shown in Fig.\ref{fig:model_schematic}. 
There are 3 main components in this model: an encoder, a decoder, and a translation network.
In pretraining, the encoder and the decoder are trained with masked image reconstruction task. 
The input images are patchified and masked before they are passed to the encoder. 
Tube masking \cite{wei2023maskedfeaturepredictionselfsupervised} is adopted as the masking strategy in the pipeline to help the model capture useful spatial-temporal structures. 
The encoder learns how to represent the sequence of input masked precipitation maps in the latent space, while the decoder learns how to reconstruct the images given the latent representation. 
Translation network is not involved in the pretraining.
In fine-tuning, the learning objective is to predict future precipitation maps given the past ones. 
The translation network aims to translate the latent representation of the past images generated by the encoder to that of the future images, so that the pretrained decoder can reconstruct the images given the correct latent representation.

\textbf{Hierarchical Encoder and Decoder} 
We use SparK \cite{tian2023designing}, the self-supervised learning convolutional encoder-decoder model, as the foundation. 
SparK built the encoder and the decoder in a hierarchical fashion as UNet \cite{ronneberger2015u}. 

The encoder is composed of sparse convolution layers, and it downsamples the input images and generates multi-scale feature maps of different resolutions. 
We choose ResNet-18 \cite{he2016deep} (Fig.\ref{fig:model_schematic}b, left side) as the encoder unless further specified. 
Given an image with height $H$ and width $W$, the encoder generates sparse feature maps $\mathbf{S}=\{S_i\mid i=1...4\}$ at 4 scales with resolutions of $\frac{H}{4} \times \frac{W}{4}$, $\frac{H}{8} \times \frac{W}{8}$, $\frac{H}{16} \times \frac{W}{16}$, and $\frac{H}{32} \times \frac{W}{32}$ respectively. 
% \begin{equation}
%   \mathbf{S} = \{S_1,S_2,S_3,S_4\} \in \{\mathbb{R}^{\frac{W}{4}\times\frac{H}{4}},\mathbb{R}^{\frac{W}{8}\times\frac{H}{8}},\mathbb{R}^{\frac{W}{16}\times\frac{H}{16}},\mathbb{R}^{\frac{W}{32}\times\frac{H}{32}}\}
% \end{equation}
Since the feature maps generated by the encoder are sparse, a densify network is necessary to fill in the empty positions of sparse feature maps with mask embedding $[M_i]$ at the $i$-th level as follows: 
$$S_i^\prime=S_i+[M_i]\text{,}$$ 
where $[M_i]$ is a learnable 2D map parameter, initialized with truncated normal distribution as in SparK, and $S_i^\prime$ is the dense feature map at the $i$-th level. The decoder network decodes the dense feature maps and then reconstructs the original images.
We use the same \textit{LightDecoder} as in SparK .
The \textit{LightDecoder} consists of 3 successive UNet blocks (Fig.\ref{fig:model_schematic}b, right side), $\{B_1, B_2, B_3\}$ with different hyperparameters, to upsample and decode features from corresponding levels and a Conv2D layer for final projection.

Denoting the set of decoded feature maps as $\mathbf{D}=\{D_i\mid i=1...4\}$ and the set of projection layers $\mathbf{\phi}=\{\phi_i\mid i=1...4\}$ for 4 levels, composed of normalization layers and Conv2D layers. $D_4$ is then projected by projection layer $\phi_4$:
$$D_4=\phi_4(S_4^\prime)\text{,}$$
for $\{D_i \mid i=3...1\}$, the decoded feature map is derived from that of the lower level $D_{i+1}$ and the dense feature map at the same level $S_i^\prime$. 
$S_i^\prime$ is projected to comply with the decoded feature map dimension with the projection layer $\phi_i$ at the $i$-th layer:
$$D_i=B_i(D_{i+1})+\phi_i(S^\prime_i), i=3...1\text{.}$$
At last, the output prediction $\hat{X}$ is derived from the final projection on $D_1$ as follows:
$$\hat{X}=\text{Conv2D}(D_1)\text{,}$$ 
where $\hat{X}$ is the output prediction of precipitation maps depending on the learning objective in pretraining or fine-tuning phase. 

\textbf{Translation Network} 
We use a translation network to ease the task of the encoder and the decoder in fine-tuning. 
% argument-why better?
In the MIM pretraining, the encoder and decoder can collaborate well with each other to represent the input sequence in latent space and then reconstruct it. 
However, in the fine-tuning, it is difficult for the encoder and the decoder to shift from the learned knowledge since they were well-trained on the reconstruction task.
The translation network can manage to learn the shift between the past and future representations, so that the well-trained encoder and decoder can fully use their abilities of extracting and reconstructing from representations learned in pretraining.
The joining of translation network reliefs the difficulty of fine-tuning the parameters of encoder and decoder compared to the approaches without translation network, e.g. SparK.
The encoder, the translation network, and the decoder are the main components of SpaT-SparK.
The three components collaborate to make accurate predictions.

The translation network maps the latent representation of the past image sequence at the $i$-th level $S_{p,i}$ to that of the corresponding future image sequence $S_{f,i}$. $\mathbf{S}_p$ and $\mathbf{S}_f$ are the sets of features of past and predicted future precipitation maps obtained at different levels of the network respectively. 
% This relieves the difficulty of training the decoder, since decoder is well trained in the MIM pretraining. 
Since the features are 2-dimensional, the translation network are expected to take the 2D features as input and then output 2D features.
To simplify, we select the combination of a Conv2D layer and tanh layer as non-linear layer for features at 4 levels. 
We refer translation network as $\Theta$, and the sublayer at the $i$-th layer as $\theta_i$:
$$\theta_i(\cdot)=\text{tanh}(\text{Conv2D}(\cdot))\text{,}$$
% The future representation $\mathbf{S}_f$: 
$$\mathbf{S}_f=\Theta(\mathbf{S}_p)=\{S_{f,i}\mid S_{f,i}=\theta_i({S_{p,i}}), i=1...4\}\text{.}$$

\textbf{Pretraining and Fine-tuning}
In pretraining, the objective of SpaT-SparK is to reconstruct the precipitation map sequence of length $T$.
SpaT-SparK model leverage the advantages of learning hierarchical and sparse feature representation in SparK. 
The optimization goal is to minimize the per-patch normalized L2 error between the prediction and the ground truth precipitation maps on masked patches only.
The pretraining can be abstracted from a high level as:
$$\underset{\mathbf{W}}{\arg\min} \text{ loss}({X}_{t\rightarrow t+T-1}\ast mask, \hat{X}_{t\rightarrow t+T-1} \ast mask) \text{,}$$ 
where
$$\hat{X}_{t\rightarrow t+T-1}=\text{DEC}(\text{Densify}(\text{ENC}({X}_{t\rightarrow t+T-1})))\text{.}$$
$\mathbf{W}$ refers to the weights of encoder, decoder and densify network in SpaT-SparK model, $\hat{X}_{t\rightarrow t+T-1}$ is the reconstruction of the precipitation map sequence from time $t$ to $t+T-1$ while $X_{t\rightarrow t+T-1}$ is the input precipitation map sequence from time $t$ to $t+T-1$.

In fine-tuning, the objective of SpaT-SparK is to translate the precipitation map sequence in the past to that in the future. 
The pretrained components of SpaT-SparK model performs the major task of encoding and decoding the past and future precipitation maps, and help to train the translation network from scratch. 
The fine-tuning can be abstracted from a high level as:
$$\underset{\mathbf{W}, \mathbf{W}_{T}}{\arg\min} \text{ loss}({X}_{t+T\rightarrow t+2T-1}, \hat{X}_{t+T\rightarrow t+2T-1}) \text{,}$$
where
\begin{equation*}
  \begin{split}
    \hat{X}_{t+T\rightarrow t+2T-1}&=\\\text{DEC}(&\text{Densify}(\text{Translate}(\text{ENC}({X}_{t\rightarrow t+T-1}))))\text{.}
  \end{split}
\end{equation*}
$\mathbf{W}_{T}$ refers to the weight of the translation network of SpaT-SparK model.

\section{NL-50 Dataset}

The NL-50 dataset \cite{trebing2021smaatunet} is a precipitation map dataset within the range of the Netherlands and its surroundings from 2016-2019, measured by Royal Netherlands Meteorological Institute\footnote[1]{https://dataplatform.knmi.nl/} (KNMI, Koninklijk Nederlands Meteorologisch Instituut). 
% Radar reflectivities are measured by both radars (De Bilt at 52.1017 N, 5.1783 E and Den Helder at 52.9533 N, 4.7900 E) in scans with elevation angles of 0.3, 0.4, 0.8, 1.1, 2.0, 3.0, 4.5, 6.0, 8.0, 10.0, 12.0, 15.0, 20.0, and 25.0 degrees. These scans are used to compute echo top heights by taking the maximum height above the earth?s surface where reflectivity exceeds 7 dBZ. These two radar images are then combined to single echo top height values by taking a the maximum of the two echo top heights. The time stamp in the file name corresponds to the start time of the lowest radar scan.
The data is measured by both radars in De Bilt (52.1017 N, 5.1783 E) and Den Helder (52.9533 N, 4.7900 E). 
We followed the same data preprocessing steps as in \cite{trebing2021smaatunet} and also use min-max scaled dataset. Similar to \cite{trebing2021smaatunet}, we filter the data where the number of above-threshold pixels are greater than certain values. 
We set the threshold to  0.5mm/h, and select the map images where the number of above-threshold pixels greater than 50\% of single the pixels. 
We define the length of input image sequences for SpaT-SparK model as $T$.
Then we combine the first $T$ maps with subsequent $T$ maps to make sequences: the first $T$ maps are used as input images for both MIM pretraining and fine-tuning, and the last $T$ maps are used as prediction target for fine-tuning. 
In this work, $T$ images are converted to a sequence, therefore the input and the output both have the shape of $[T,H,W]$.
The dataset is partitioned to training (2016-2018) and testing sets (2019) by year. 
The training set has 5734 datapoints while the testing set has 1557 datapoints. 

\section{NL-50 Experiment}

% \begin{itemize}
%   \item performance compare table \ref{tab_performance_comparison}
%   \item performance on individual time figure \ref{fig:single_time_performance_all}
% \end{itemize}

\subsection{Experiment Settings}
We choose different experiment settings for pretraining and fine-tuning phases. 
We use sequences length $T=12$ in both pretraining and fine-tuning. In pretraining, we choose batch size 256, base learning rate $2\times10^{-4}$ with linear learning rate annealing scheduler, training epoch 1400, warm-up epoch 40, mask ratio 0.6 as the hyperparameters, and LAMB optimizer \cite{you2019large} with weight decay 0.04 to update the weights. 
The optimization objective is to minimize the per-patch normalized L2 error on the masked patches. 
This has been proven to enable the model to learning more informative features from data 
\cite{he2022masked,tian2023designing}. In the fine-tuning, we choose batch size 196, base learning rate $1.5\times10^{-4}$, training epoch 200, warm-up epoch 40, mask ratio 0.6 as the hyperparameters, and LAMB optimizer with weight decay 0.05 to update the weights. 
The optimization objective is to minimize the L2 error on both masked and unmasked patches.
% Table \ref{tab:hparams}

\subsection{Metrics} \label{sec:metrics}

We evaluate the model performances with multiple metrics.
To calculate the metrics, the raw output and the model prediction are reverted with inverse min-max scaling (unit: mm/5min). 
The ground truth and prediction maps are thresholded with 0.5mm/h, and pixel-wise positive and negative masks are created. 
Given those, we compute the confusion matrix and mark true positive (TP), false positive (FP), false negative (FN), and true negative (TN) pixels.
Similar to \cite{trebing2021smaatunet}, we use pixel-wise MSE (pMSE, eq.\ref{eq:pMSE}, $\hat{X}, X$ are predictions and ground truth respectively  ), accuracy, precision, recall, F1 score, critical success index (CSI, eq.\ref{eq:csi}), false alarm rate (FAR, eq.\ref{eq:far}), and Heidke skill score (HSS, eq.\ref{eq:hss}) to evaluate the performance of the evaluated models. 
{
  \small
  \begin{align}
    & \text{pMSE} = \frac{\Vert X-\hat{X}\Vert_2^2}{H\times W} \label{eq:pMSE}\text{,} \\
    & \text{CSI} = \frac{TP}{TP+FP+FN} \label{eq:csi}\text{,} \\
    & \text{FAR} = \frac{FP}{TP+FP} \label{eq:far}\text{,} \\ 
    & \text{HSS} = \frac{TP\times TN - FP\times FN}{(TP+FN)(FN+TN)+(TP+FP)(FP+TN)} \text{.} \label{eq:hss}
  \end{align}
}% 

\subsection{Results}

\textbf{Overall Performance}
We compared the performances of SpaT-SparK and the baseline models, i.e. SmaAt-UNet and SparK with ResNet-18 or ResNet-50 as the encoder, in Table \ref{tab_performance_comparison}. The results showed that SpaT-SparK with ResNet-18 as the encoder has the lowest pMSE. 
Comparing to SmaAt-UNet, the pMSE decreased by $8.9\%$.
Comparing to SparK, the pMSE decreased in either encoder ($-1\times10^{-4}$ and $-7\times10^{-4}$ for ResNet-50 and ResNet-18, respectively). 
This decrease indicates that the translation network is an effective component to perform the temporal transformation. 
With the collaboration between the translation network and the encoder and decoder, the learning task of the encoder and decoder is less difficult than that in SparK model.
Notably, SpaT-SparK achieves the highest accuracy among all the models, outperforming the baseline SmaAt-UNet model by large margins ($18\%$). 
The precision metric of the SpaT-SparK model improved by $7.4\%$, and FAR improved by $12.8\%$. 
However, there are drops in recall, F1, CSI, and HSS.
The results shine some light on that SSL can make convolutional networks more powerful in precipitation nowcasting.

% TODO Mention visualization
% 
% ! OVER !
\begin{table*}[htbp] % table performances over single
  \centering
  \caption{The performance comparison among SmaAt-UNet, SparK and SpaT-SparK with ResNet-18 or ResNet-50 as the encoder. ``$\uparrow$" indicates the higher the value is, the better the performance is. ``$\downarrow$" indicates the lower the value is, the better the performance is. The best result in each column is marked as bold. }
  \begin{tabular}{l c c c c c c c c r}
    \toprule
    Model & pMSE $\downarrow$ & Accuracy $\uparrow$ & Precision $\uparrow$ & Recall $\uparrow$ & F1 $\uparrow$ & CSI $\uparrow$ & FAR $\downarrow$ & HSS $\uparrow$ \\
    \hline
    SmaAt-UNet & 0.0145 & 0.774 & 0.631 & \textbf{0.846} & \textbf{0.723} & \textbf{0.566} & 0.368 & \textbf{0.269} \\ 
    SparK\\
    \hspace{0.3cm} \textit{ResNet-50} & 0.0136 & 0.911 & 0.646 & 0.520 & 0.543 & 0.373 & 0.353 & 0.245 \\ 
    \hspace{0.3cm} \textit{ResNet-18} & 0.0139 & 0.910 & 0.620 & 0.444 & 0.512 & 0.344 & 0.379 & 0.232 \\
    SpaT-SparK \\ 
    \hspace{0.3cm} \textit{ResNet-50} & 0.0135 & 0.911 & 0.633 & 0.486 & 0.528 & 0.359 & 0.366 & 0.239 \\
    \hspace{0.3cm} \textit{ResNet-18} & \textbf{0.0132} & \textbf{0.913} & \textbf{0.678} & 0.588 & 0.560 & 0.389 & \textbf{0.321} & 0.255 \\
    % SparK-LinearB & 0.0138 & 0.911 & 0.640 & 0.528 & 0.547 & 0.377 & 0.360 & \\
    \bottomrule
  \end{tabular}
  \label{tab_performance_comparison}
  \end{table*}

\begin{table*}[htbp] % table ablation study
  \centering
  \caption{Ablation study results. ``$\uparrow$" indicates the higher the value is, the better the performance is. ``$\downarrow$" indicates the lower the value is, the better the performance is. The best result in each column is marked as bold. }
    \begin{tabular}{l c c c c c c c c r}
      \toprule 
      Method & pMSE $\downarrow$ & Accuracy $\uparrow$ & Precision $\uparrow$ & Recall $\uparrow$ & F1 $\uparrow$ & CSI $\uparrow$ & FAR $\downarrow$ & HSS $\uparrow$ \\
      \hline
      SpaT-SparK(ResNet18) & \textbf{0.0132} & \textbf{0.913} & 0.678 & 0.588 & \textbf{0.560} & \textbf{0.389} & 0.321 & \textbf{0.255} \\ 
      w/o pretrain & 0.0147 & 0.906 & 0.623 & 0.469 & 0.498 & 0.331 & 0.377 & 0.220 \\
      frozen- \\ 
      \hspace{0.3cm} encoder & 0.0140 & 0.912 & \textbf{0.687} & \textbf{0.602} & 0.545 & 0.375 & \textbf{0.313} & 0.246 \\
      \hspace{0.3cm} decoder & 0.0144 & 0.911 & 0.679 & 0.460 & 0.510 & 0.343 & 0.320 & 0.230 \\
      \hspace{0.3cm} encoder+decoder & 0.0147 & 0.909 & 0.642 & 0.444 & 0.495 & 0.329 & 0.358 & 0.221 \\
      \bottomrule
    \end{tabular}
  \label{tab:abaltion_study}
  \end{table*} 

\begin{figure}[htbp] % figure performance on single time
  \begin{tabular}{cc}
    \begin{subfigure}[c]{0.5\columnwidth}
      \caption{pMSE $\downarrow$}
      \includegraphics[width=\columnwidth]{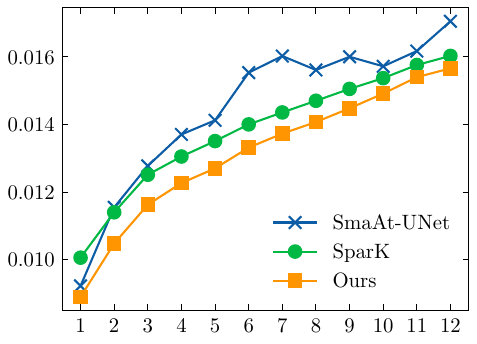}
      \label{fig:singletime_loss}
    \end{subfigure} &
    \begin{subfigure}[c]{0.5\columnwidth}
      \caption{Accuracy}
      \includegraphics[width=\columnwidth]{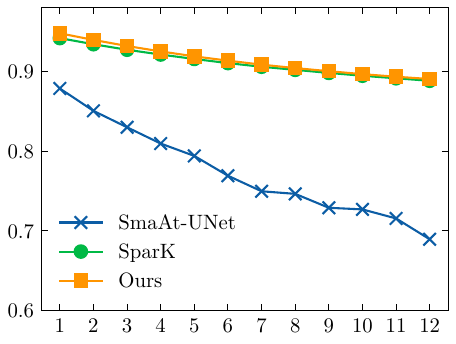}
      \label{fig:singletime_acc}
    \end{subfigure}
  \end{tabular}
  \begin{tabular}{cc}
    \begin{subfigure}[c]{0.5\columnwidth}
      \caption{Precision}
      \label{fig:singletime_precision}
      \includegraphics[width=\columnwidth]{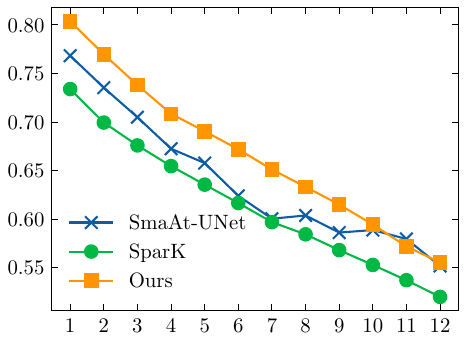}
    \end{subfigure} &
    \begin{subfigure}[c]{0.5\columnwidth}
      \caption{Recall}
      \includegraphics[width=\columnwidth]{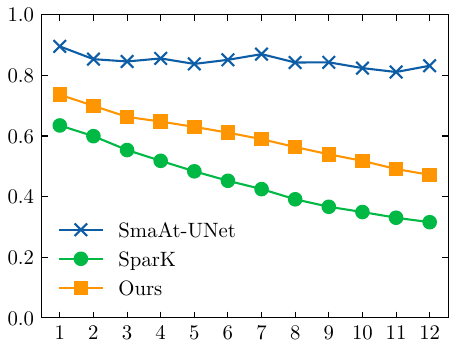}
      \label{fig:singletime_recsingle} 
    \end{subfigure} 
  \end{tabular}
  \begin{tabular}{cc}
    \begin{subfigure}[c]{0.5\columnwidth}
      \caption{F1}
      \includegraphics[width=\columnwidth]{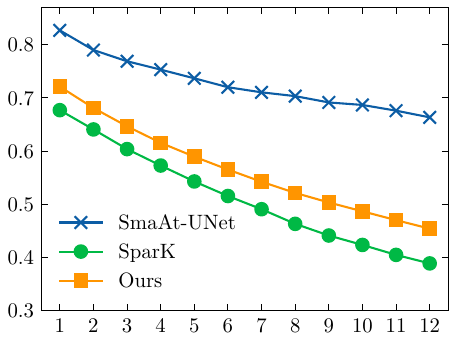}
      \label{fig:singletime_f1}
    \end{subfigure} & 
    \begin{subfigure}[c]{0.5\columnwidth}
      \caption{CSI}
      \includegraphics[width=\columnwidth]{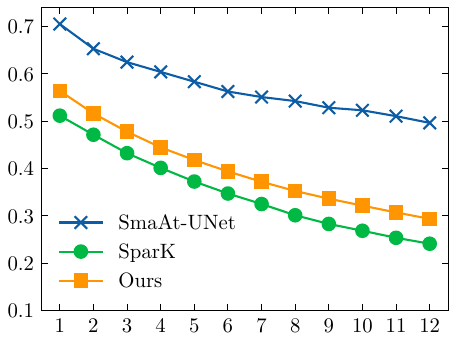}
      \label{fig:singletime_csi}
    \end{subfigure} 
  \end{tabular}
  \begin{tabular}{c c}
    \begin{subfigure}[c]{0.5\columnwidth}
      \caption{FAR $\downarrow$}
      \includegraphics[width=\columnwidth]{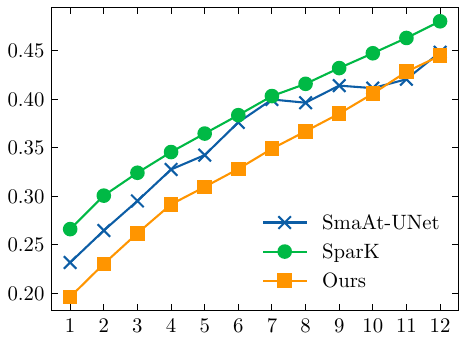}
      \label{fig:singletime_far}
    \end{subfigure} &
    \begin{subfigure}[c]{0.5\columnwidth}
      \caption{HSS}
      \includegraphics[width=\columnwidth]{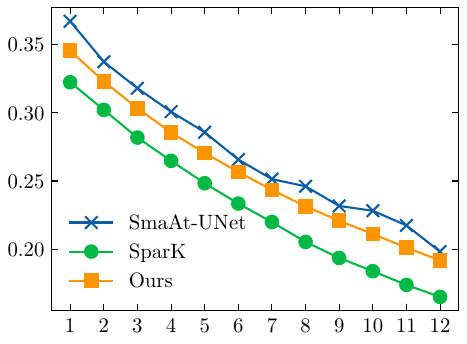}
      \label{fig:singletime_hss}
    \end{subfigure} 
  \end{tabular}
  \caption{Performance of the models at each time step. SparK and SpaT-SparK both use ResNet-18 as the encoder. "$\downarrow$" indicates the lower the value is, the better the performance is.}
  \label{fig:single_time_performance_all}
\end{figure}

\textbf{Performance on Single Time Step}
We evaluated the model performance on each single time step to investigate the reliability of model with different time offsets. 
The metrics used for overall performance elevation are also used for evaluations on each single time step.
The time offset is referred as $t$, where $t=1$ indicates the first prediction map. 
The results are shown in Fig.\ref{fig:single_time_performance_all}. In Fig.\ref{fig:singletime_loss}, we can observe that pMSE increased as $t$ increases for all models. 
SpaT-SparK showed the lowest pMSE among all models from $t$=1. 
This shows that SpaT-SparK is more accurate than others in the 1-hour period precipitation nowcasting. 
In Fig.\ref{fig:singletime_acc}, for accuracy, we can observe that SpaT-SparK outperforms SmaAt-UNet by a large margin and it has a moderate decrease as $t$ increases. 
In Fig.\ref{fig:singletime_precision} and \ref{fig:singletime_far}, we can observe that SparK underperforms SmaAt-UNet but our method outperforms SmaAt-UNet. This indicates that the translation network can help to improve precision and FAR performance. The improvements by the translation network can also be observed in Fig.\ref{fig:singletime_recsingle} to \ref{fig:singletime_csi} and \ref{fig:singletime_hss} for recall, F1, CSI, and HSS, respectively . This shows that the translation network is beneficial for encoder-decoder structured spatial-temporal model in precipitation nowcasting.

\begin{remark}
  While the results showed that SpaT-SparK outperforms baseline models in some numerical metrics,
  the visualizations of the predictions did not show sufficient quality improvements. 
  % We observed that SpaT-SparK tended to predict with higher values and missed some temporal patterns in the predictions. 
  % Also we noticed that SpaT-SparK produced blurred predictions compared to the ground truth. 
  This is probably the result of using a relatively large patch-size and the capability limitation of the decoder for generation.
\end{remark}

\textbf{Efficiency} 
The inference time is another important factor for precipitation nowcasting applications. 
A method with shorter inference time and more accurate predictions are preferred.
We measured the FLOPs, model size, and the model inference time for a single input on 1$\times$ NVIDIA A100 80GB GPU.
The results are shown in Table \ref{tab:efficiency}. 
% The SpaT-SparK model can infer the future precipitation map in 1.340s, even though introducing the ResNet as encoder and the translation network. 
We think that keeping the inference time in the magnitude of seconds makes it feasible as an application.
\begin{table}[htbp]
  \centering
  \caption{Efficiency performance. The data is measured on 1$\times$ NVIDIA A100 80GB GPU. }
  \begin{tabular}{l c c c c}
    \toprule
    Model & FLOPS(G) & \#params(M) & pMSE \\ 
    \hline
    SmaAt-UNet & 24.7 & 4.0 & 0.0145 \\
    SparK \\ 
    \hspace{0.3cm} \textit{ResNet-18} & 99.2 & 43.4 & 0.0139 \\
    \hspace{0.3cm} \textit{ResNet-50} & 110.8 & 53.0 & 0.0136 \\
    SpaT-SparK \\ 
    \hspace{0.3cm} \textit{ResNet-18} & 172.6 & 96.5 & 0.0132 \\
    \hspace{0.3cm} \textit{ResNet-50} & 184.1 & 106.0 & 0.0135 \\
    \bottomrule
  \end{tabular}
  \label{tab:efficiency}
\end{table}

\subsection{Ablation Study}

We further investigate the effect of the pretraining and fine-tuning strategies. 
In Table \ref{tab:abaltion_study}, we compare the model performance with an unpretrained encoder and/or decoder, a frozen encoder or decoder during fine-tuning, and summarize the results. 
The results showed that the complete SpaT-SparK model outperformed others in pMSE, accuracy, F1-score, CSI, and HSS. 
Without self-supervised pretraining, SpaT-SparK shows an obvious drop in all performance metrics. This necessitates the introduction of self-supervised pretraining again. 
The frozen encoder during fine-tuning enabled the model to perform better than others in precision, recall and FAR, while only minor decreases are observed in accuracy, F1-score, and CSI. 
We also observed that SpaT-SparK model could achieve faster convergence to the best performance compared to those without pretraining, or partially/completely parameter freezing strategies. 
In conclusion, we think both the self-supervised pretraining and the translation network improved the performance of precipitation nowcasting.

% \subsection{Visualization}\label{sec:visualization}
% 
% Visualizations of the predicted precipitation maps by different models are shown in figure \ref{fig:visualization_secresult}.
% % TODO
% \begin{figure}[h]
%   \centering
%   \includegraphics[width=0.7\columnwidth]{example-image-a}
%   \caption{Visualization}\label{fig:visualization_secresult}
% \end{figure}
% 
\section{Conclusion and Further Work}

% Lowering the errors of predictions is the key to more accurate precipitation nowcasting. 
% End-to-end supervised training enable spatialtemporal learning models showing their power in precipitation nowcasting applications. 
% Self-supervised learning methods benefit the progress in various computer vision tasks. 
% With self-supervised learning, we can imagine a boost towards more accurate precipitation nowcasting. 

In this study, we proposed SpaT-SparK model to address this problem by leveraging the power of self-supervised learning and spatial-temporal modelling. 
We pretrained the encoder and the decoder in a self-supervised masked image modeling fashion.
Then we employ a translation network to tackle the spatial-temporal transformation problem in fine-tuning. 
Our study reveals the potential of applying self-supervised learning in precipitation nowcasting. 
Future studies can focus on selecting optimal self-supervised learning strategy for learning representations of precipitation maps, and designing a more efficient translation network to capture capturing the dynamics of learned representations in the latent space.

\bibliographystyle{IEEEtran}
\bibliography{IEEEabrv,ref}

\end{document}